\RequirePackage[OT1]{fontenc}
\documentclass[letterpaper,10pt,conference]{ieeeconf}

\usepackage{graphicx}
\usepackage{amsmath}
\usepackage{amssymb}
\usepackage{multirow}
\usepackage{booktabs}
\let\labelindent\relax
\usepackage{enumitem}
\usepackage{cite}
\usepackage{url}
\usepackage{xcolor}
\usepackage[normalem]{ulem}                     

\usepackage{colortbl}
\usepackage[hidelinks]{hyperref}
\usepackage{balance}

\newcommand{\eg}{\emph{e.g.}}
\newcommand{\ie}{\emph{i.e.}}
\IEEEoverridecommandlockouts
\makeatletter
\long\def\@makecaption#1#2{%
  \par\vskip 5pt%
  \noindent\parbox{\linewidth}{\normalfont\small #1: #2\par}%
}
\makeatother

\newcommand{\method}{SatFix}
\newcommand{\dataset}{University-Metric}
\newcommand{\task}{Absolute Metric UAV Localization}

\title{\LARGE \bf
\method: Absolute Visual Localization of UAVs in Satellite Maps \\ from a Single Oblique Image
}

\author{Jiarui Zeng$^{1,2}$, Kun Shi$^{1,2}$, Chiman Vong$^{1,2}$ and Zhedong Zheng$^{1,2}$%
\thanks{$^{1}$Jiarui Zeng, Kun Shi, Chiman Vong and Zhedong Zheng are with the Department of Computer and Information Science,
        Faculty of Information Science and Computing,
        University of Macau, Macao SAR, China.}%
\thanks{$^{2}$Jiarui Zeng, Kun Shi, Chiman Vong and Zhedong Zheng are with the Institute of Artificial Intelligence and Brain Sciences,
        University of Macau, Macao SAR, China. Corresponding author is Zhedong Zheng
        ({\tt\small zhedongzheng@um.edu.mo}).}
}

\begin{document}
\raggedbottom

\maketitle
\thispagestyle{empty}
\pagestyle{empty}

\begin{abstract}
We study absolute metric UAV localization within a provided geo-referenced
satellite region: recovering continuous map position and viewing heading from
a single oblique image, or a short clip of multiple views. Existing cross-view geo-localization methods retrieve the most similar satellite tile from a gallery and report Recall@$K$. However, retrieval depends on gallery sampling, carries no heading
estimate, and returns a tile index rather than a continuous coordinate. To this end, we propose \method{}, a feed-forward UAV--satellite localization framework built on VGGT-$\Omega$: satellite-grid features act as queries that aggregate UAV visual evidence, and two lightweight heads regress a 3-DoF pose in the satellite-map frame,
\ie, continuous 2D position and heading. Unlike reconstruction-based pipelines, \method{} needs no explicit 3D map, rendered bird's-eye image, or auxiliary sensor. A single model supports single- and multi-view inputs,
with trajectory constraints applied during multi-view training. To enable metric evaluation, we introduce \dataset{}, in which satellite imagery is re-collected over a region up to $10.7\times$ longer on a side ($\approx$ $114\times$ the ground area) than the original University-1652 tiles, with continuous position and heading labels of the original UAV tours. With one UAV view, \method{} localizes $52.08\%$ of test frames
within 50~m and $17.34\%$ within 10~m, with a median
position error of 45.66~m and a median heading error of $20.81^\circ$. Inference takes less than 0.1~seconds per single-view query on an
NVIDIA RTX 4090; \method{} directly predicts in the geo-referenced
satellite-map frame without test-time pose alignment. In the controlled multi-view evaluation, using nine UAV views further
reduces the median position error to 21.96~m and the median heading error
to $8.73^\circ$. Against a fine-tuned VGGT-$\Omega$ baseline, \method{} lowers the median position error by $34.0\%$ and the nine-view median heading error from $25.43^\circ$ to $8.73^\circ$.  
\end{abstract}

\section{INTRODUCTION}
\label{sec:intro}

\begin{figure}[t]
    \centering
    \includegraphics[width=\linewidth]{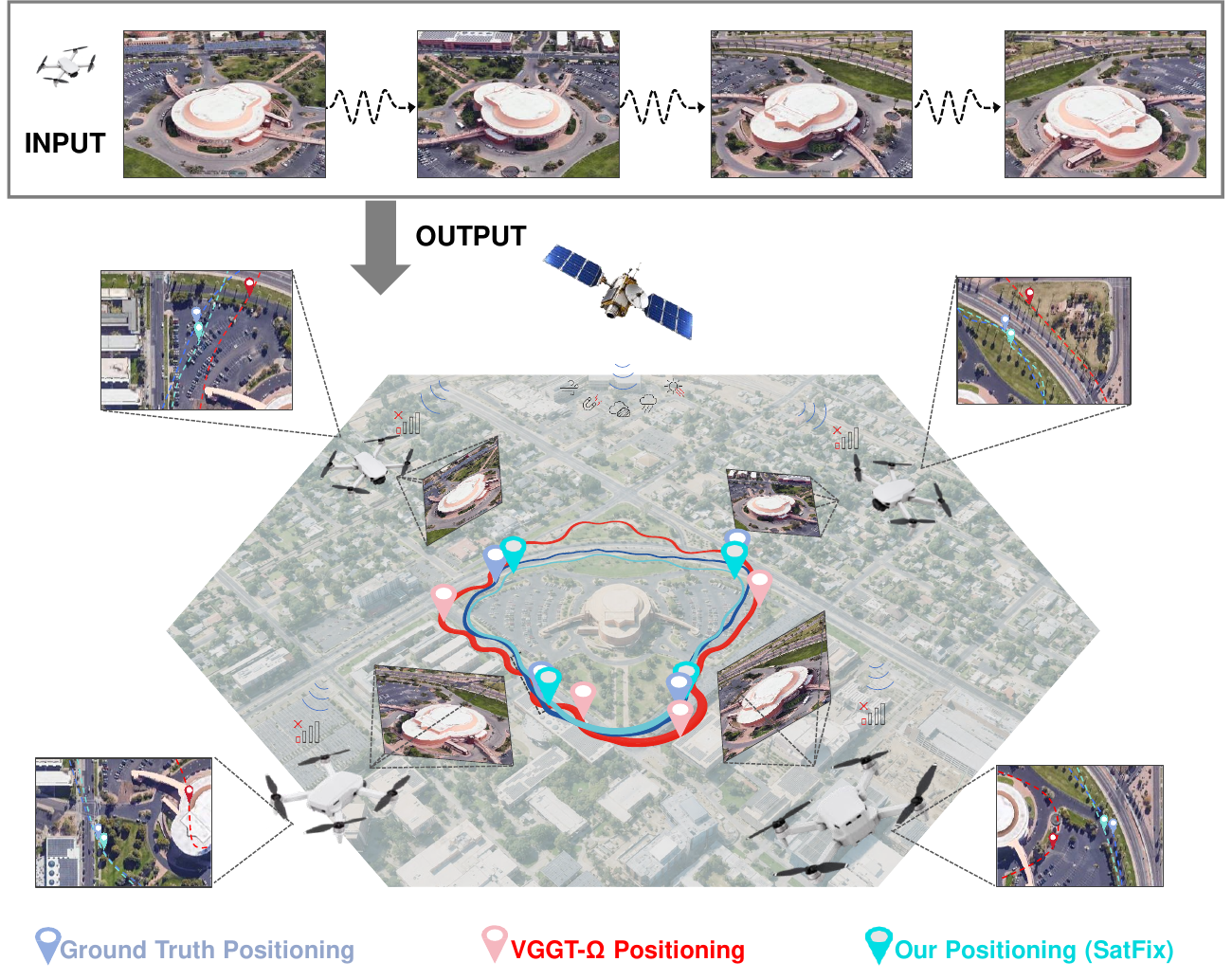}
    \vspace{-.2in}
    \caption{\textbf{Task motivation.} When GNSS is unavailable, \eg, in urban canyons or under dense canopy, absolute metric UAV localization can use onboard vision.
Given a single UAV view or a short clip of multiple views as input,
the model outputs a position and a heading on a geo-referenced
satellite map. Top: the UAV clip input and the satellite map. Bottom: ground-truth (\textcolor{blue}{blue}), VGGT-$\Omega$ zero-shot (\textcolor{red}{red}), and \method{} (\textcolor{cyan}{cyan}) trajectories.}
    \label{fig:task_overview}
    \vspace{-.2in}
\end{figure}

Given the satellite map and what the camera sees, a skilled operator
can infer the UAV position. In this work, we study
\textbf{\task} based on visual inputs. Given either a single oblique UAV image or a short clip of several views, plus one geo-referenced RGB satellite crop of a known operating region, we aim to recover the continuous position and viewing heading of the UAV camera in the satellite map (see Fig.~\ref{fig:task_overview}). Yet we face two challenges: (1) existing benchmarks rarely combine oblique
UAV views with a large regional satellite reference; and (2) efficient
feed-forward methods for this setting remain under-explored.

For the first challenge, most existing benchmarks cast localization as retrieval from a gallery of geo-tagged satellite tiles~\cite{zheng2020university1652,dai2024denseuav,deuser2023sample4geo} and report Recall@$K$. The satellite references of these benchmarks span only ${\sim}118$--$120$~m on a side (Table~\ref{tab:benchmark_comparison}); a tile index encodes neither
a continuous position nor a heading. 
Other benchmarks~\cite{xu2024uavvisloc,li2025geovins,he2024aerialvl} do cover a larger extent, but these datasets acquire the UAV imagery at a near-nadir view, where a top-down frame and the reference crop show nearly the same scene; the task then loses much of the cross-view difficulty associated with
forward-looking oblique UAV observations. We therefore study regional localization under a coarse geographic prior,
where the UAV is known to lie within a provided geo-referenced satellite
region rather than an unconstrained global search space.
To support metric evaluation at region scale with oblique views, we introduce
\dataset{}, a benchmark derived from the University-1652 UAV tours with a
location-level train/validation/test split. We re-collect the satellite map over a region up to $10.7\times$ longer on a side ($\approx 114\times$ the ground area) than the original per-building crops and inherit the tours' continuous position and heading labels. The benchmark contains 83{,}160 UAV frames from 1{,}540 locations, split by location into 631 training, 70 validation, and 839 test locations.  

For the second challenge, we propose \method{}, a feed-forward framework that anchors geometric representations to the satellite coordinate frame. Specifically, satellite-grid features serve as queries and UAV features serve as keys and values in cross-attention to estimate continuous position and heading. A single model supports both single-view and multi-view inputs. In summary, our contributions are:  
\begin{itemize}[leftmargin=5pt]
\item \textbf{Region-scale oblique benchmark.} Existing geo-localization benchmarks retrieve from a gallery whose tiles span only ${\sim}118$~m on a side, or capture near-nadir frames that nearly duplicate the reference crop. We introduce \dataset{}, which pairs oblique UAV imagery with a regional satellite map whose default crop is $5.3\times$ longer on a side than University-1652 tiles and whose largest crop is $10.7\times$ longer, and per-frame position--heading labels, split by location.  
\item \textbf{Grounding by cross-attention.} We propose \method{}, which forgoes scene reconstruction and tile retrieval: satellite-grid features query UAV features through a single cross-attention, and two light heads decode a 3-DoF pose on top of a pretrained VGGT-$\Omega$ backbone, without a 3D map, auxiliary sensor, or test-time alignment.
\item \textbf{Accuracy and efficiency.} Our \method{} localizes $52.08\%$ of single-view frames within $50$~m (median $45.66$~m), and, with nine views, reaches $21.96$~m median position and $8.73^\circ$ median heading, all in one forward pass under $0.1$~s per query on a single RTX 4090, faster than WildNav and Reloc3r.  
\end{itemize}

\begin{table*}[t]
\centering
\setlength{\tabcolsep}{3pt}
\renewcommand{\arraystretch}{1.15}
\footnotesize
\resizebox{\textwidth}{!}{%
\begin{tabular}{@{}lccccccc@{}}
\toprule
\textbf{Benchmark} &
\textbf{Input} &
\textbf{\# UAV Imgs.} &
\textbf{Sat. Extent (m)} &
\textbf{UAV Viewpoint} &
\textbf{Sat. Ref.} &
\textbf{Loc. Label} &
\textbf{Task} \\
\midrule
University-1652~\cite{zheng2020university1652}
& Single
& 89.2K
& 118
& Oblique
& Gallery
& ID
& Retrieval \\
SUES-200~\cite{zhu2023sues200}
& Single
& 40.0K
& N/A
& Oblique
& Gallery
& ID
& Retrieval \\
DenseUAV~\cite{dai2024denseuav}
& Single
& 9.1K
& 120
& Nadir
& Gallery
& Coord.
& Retrieval \\
UAV-VisLoc~\cite{xu2024uavvisloc}
& Single
& 6.7K
& 11000
& Nadir
& Regional map
& Coord.+Heading
& Localization \\
GeoVINS~\cite{li2025geovins}
& Sequence
& 18,361
& 50000
& Nadir
& Regional map
& Coord.+Heading
& Localization \\
AerialVL~\cite{he2024aerialvl}
& Single/Sequence
& 18.4K
& 300
& Nadir
& Regional map
& Coord.
& Retrieval/Localization \\
University-Pose~\cite{li2026unicvgl}
& Single/Sequence
& N/A
& 118
& Oblique
& Gallery
& Coord.+Heading
& Retrieval/Localization \\
KoSim-GL~\cite{ahn2026kosimgl}
& Single/Sequence
& 2.45M
& N/A
& Nadir/Oblique
& Gallery
& ID+Coord.
& Retrieval \\
\midrule
\dataset{} (Ours)
& Single/Sequence
& 83.2K
& \textbf{473/630/1261}
& \textbf{Oblique}
& Regional map
& Coord.+Heading
& Localization \\
\bottomrule
\end{tabular}%
}
\caption{Comparison of UAV geo-localization benchmarks and released
evaluation data. Sat. Extent reports approximate ground side lengths of
satellite references in meters; regional-map extents may vary by
location. The AerialVL image count refers to its visual place
recognition subset. GeoVINS denotes its released evaluation data.
N/A denotes information not reported in the corresponding source.}   
\label{tab:benchmark_comparison}
\vspace{-.15in}
\end{table*}
\section{RELATED WORK}
\label{sec:related}
\noindent\textbf{Retrieval-based Cross-view Geo-localization.}
Cross-view geo-localization commonly matches a query image against a geo-tagged aerial or satellite gallery. Recent methods improve retrieval through compressed map representations \cite{cai2024voloc}, multimodal fusion \cite{garcia2024umf}, local feature aggregation \cite{lin2022joint,song2024globalizing}, temporal consistency \cite{suomela2024placenav}, and semantic mapping \cite{kassab2024lexis}. In the UAV domain, University-1652 establishes drone-satellite cross-view retrieval \cite{zheng2020university1652} and is extended to varying weather conditions~\cite{wang2024multiple}, while SUES-200 and DenseUAV further study altitude variation and densely sampled low-altitude positioning \cite{zhu2023sues200,dai2024denseuav}. Sample4Geo, MFRGN, and DMNIL improve hard-negative learning, multi-scale representation, and label-efficient training, while more recent extensions incorporate language guidance, adverse-weather robustness, scale adaptation, and video reasoning \cite{deuser2023sample4geo,wang2024mfrgn,chen2025dmnil,chu2024geotext1652,wen2025weatherprompt,chen2026scaleadaptive,ju2025video2bev}. Despite these advances, the final geographic estimate remains associated with a retrieved gallery item and therefore depends on gallery sampling density. 

\noindent\textbf{Metric UAV Geo-localization.}
A complementary line of work moves beyond image-level retrieval toward explicit metric localization. Recent approaches study satellite-assisted matching, UAV geo-localization, satellite--thermal alignment, and visual relative localization for aerial swarms \cite{guo2024satellite,zhang2025apabi,zhou2025jrngeo,xiao2025uasthn,krizek2024swarm}. Other map-based localization systems exploit active viewpoint selection, LiDAR heat-map alignment, event--depth registration, or lightweight topo-metric maps \cite{hanlon2024activevl,wu2024lhmaploc,chen2025evloc,jiao2025litevloc}. Bearing-UAV jointly estimates position and heading from unaligned UAV--satellite observations \cite{liu2026bearing}, while AnyVisLoc studies low-altitude multi-view localization \cite{ye2026anyvisloc} and WildNav targets GNSS-free localization against open-source maps \cite{gurgu2022wildnav}. UAV-VisLoc and AerialVL further provide regional maps, frame-level geographic annotations, or aerial sequences for map-based evaluation \cite{xu2024uavvisloc,he2024aerialvl}. In contrast, we take a provided geo-referenced RGB satellite region together with one or a few oblique UAV views to predict a continuous map-frame $(x,y,\theta)$ directly, without relying on a dense 3D map or long-horizon tracking. Because the reference region is geo-referenced, position errors are reported
in meters in the metric frame of the satellite map rather than in a learned
or up-to-scale coordinate system.

\noindent\textbf{Geometry-aware Visual Localization.}
Recent visual localization methods leverage uncertainty-aware neural radiance fields, Gaussian-splatting pose refinement, compact neural implicit maps, geometry-aware local features, and efficient hierarchical retrieval \cite{chen2024nerfvl,cheng2025logs,niu2025hgsloc,zhai2025neuraloc,liu2025liftfeat,liu2025airhloc}. VGGT and VGGT-$\Omega$ learn feed-forward camera and scene geometry from a variable number of views \cite{wang2025vggt,wang2026vggtomega}, while Reloc3r directly targets generalizable relative camera pose estimation \cite{dong2025reloc3r}. University-Pose (Uni-CVGL) reconstructs a local UAV scene before satellite retrieval and pose refinement \cite{li2026unicvgl}. In contrast, \method{} directly grounds geometric features in the provided satellite coordinate frame, avoiding explicit scene reconstruction and candidate-map retrieval. CloudTrack \cite{blei2025cloudtrack} and CGTrack \cite{li2025cgtrack} address the distinct task of UAV target tracking rather than absolute UAV localization.  

\section{\dataset{}: Dataset and Task Setting}
\label{sec:dataset_setting}

\noindent\textbf{Benchmark setting and task.}
We construct \dataset{} from the UAV tours and trajectory/KML
metadata of University-1652 \cite{zheng2020university1652}. The UAV views
are rendered from Google Earth imagery following the corresponding KML camera
trajectories, while the regional satellite references are re-collected at
larger spatial extents. The largest satellite crops cover approximately $114\times$ the ground area
of the original per-building tiles (the default crop used in all main
experiments spans $630$~m on a side, \ie, $28\times$ the ground area of the
original tiles). As summarized in
Table~\ref{tab:benchmark_comparison}, earlier benchmarks primarily formulate UAV
geo-localization as gallery retrieval
\cite{zheng2020university1652,zhu2023sues200,dai2024denseuav}, whereas more
recent datasets provide regional maps and frame-level geographic annotations
for direct localization \cite{xu2024uavvisloc,he2024aerialvl}.
\dataset{} provides continuous position and heading labels for every
UAV frame, enabling metric localization from either a single UAV
view or a short clip. We will publicly release the complete \dataset{} benchmark, including
satellite imagery, frame-level annotations, and the official
train/validation/test splits. 

Let $I^s$ denote a geo-referenced satellite map and $I_i^d$
the $i$-th UAV image. Given one or more UAV images, we formulate
the task as
\begin{equation}
\left(I^s,\{I_i^d\}_{i=1}^{T}\right)
\rightarrow
\left\{
(\hat x_i,\hat y_i,\hat\theta_i)
\right\}_{i=1}^{T}.
\label{eq:task}
\end{equation}
Each drone frame has its own absolute position and heading target. Unlike gallery
retrieval, which returns a discrete reference item, our task predicts a
continuous coordinate within the provided geo-referenced region. Here,
\emph{absolute} refers to the satellite coordinate frame rather than an
unbounded geographic search space, and $T$ is the number of UAV views, \ie,
UAV images with their own camera poses ($T=1$
recovers the single-view setting). The operating region is assumed to be specified by a coarse geographic prior.
In our benchmark, this prior defines a fixed satellite-map center for each
location, while the map extent is treated as a controlled variable
(Table~\ref{tab:satellite_extent}). We
denote the ground-truth and predicted positions in the resized satellite crop by
$\mathbf p_i=(x_i,y_i)^\top$ and
$\hat{\mathbf p}_i=(\hat x_i,\hat y_i)^\top$, respectively.  

\noindent\textbf{Annotations and coordinate representation.}
Each location contains 54 UAV images sampled every $0.5$~s, while the
trajectory records are sampled every $0.1$~s. Each UAV frame corresponds
directly to a trajectory record at the same timestamp, so no temporal
interpolation or nearest-neighbor timestamp matching is required. The camera positions are then projected into the satellite coordinate frame,
while the viewing heading is taken from the corresponding KML
\texttt{LookAt} metadata. Specifically, the KML \texttt{heading} field
specifies the horizontal viewing azimuth of the virtual UAV camera, with
$0^\circ$ pointing North and angles increasing clockwise.
Throughout this paper, \emph{heading} refers to this horizontal camera-viewing
azimuth rather than the direction of motion along the trajectory. We split \dataset{} by source location identity into 631 training, 70
validation, and 839 test locations, corresponding to 34,074, 3,780, and
45,306 UAV frames, respectively. No UAV tour or source location identity is
shared across the three splits. For each location, we construct a single $1024\times1024$ satellite crop
centered at the geo-referenced map center specified by the corresponding
KML metadata. The same satellite crop is shared by all 54 UAV frames from
that location and is not recentered for individual frames.
For each location, the metric scale is derived from the corresponding
satellite KML geometry. Specifically, the ground width is computed from the
KML \texttt{range} and horizontal field of view, and divided by the original
satellite-image width to obtain meters per pixel. In the default setting,
this is approximately $0.616$~m/pixel, corresponding to a
$1024\times1024$ crop of about $630$~m on a side. We resize satellite and
UAV images to $512\times512$, and we sample short clips within the same
location.
We map the geographic positions into the resized $512\times512$ satellite-crop
pixel coordinate system using the tile's geo-referencing transform. The heading is defined as a bearing in the local horizontal plane rather
than as an image-plane direction. We order the two vector components as
$(\emph{North},\emph{East})$ and encode the heading as
$\mathbf h_i=(\cos\theta_i,\sin\theta_i)^\top=(h_{\rm N},h_{\rm E})^\top$,
where the two components are the north and east direction cosines.
This unit-vector encoding is continuous across the
$0^\circ$/$360^\circ$ boundary. Angles are converted to radians for the
trigonometric encoding. 

\noindent\textbf{Evaluation protocol and satellite extent.}
For each test frame, we map the predicted and ground-truth positions back to
the original satellite crop, compute their Euclidean pixel distance, and
convert it to meters using the location-specific meters-per-pixel scale
derived from the corresponding satellite KML metadata. We denote the resulting metric position error by $e_i$ and
report its mean and median, together with the success rate
$\mathrm{SR}@\tau=\operatorname{mean}_i\,\mathbf 1[e_i\leq\tau]$
for $\tau\in\{10,25,50,100\}$~m.
We evaluate heading using the minimum circular angular difference.  %
Beyond the accuracy protocol, we collect the satellite reference at three physical extents while keeping the
network input resolution fixed. Square crops of $768$, $1024$, and $2048$ pixels
span approximately $473$, $630$, and $1261$~m on a side and are all resized to
$512\times512$. We first evaluate the model trained on $1024$-pixel crops
at all three crop sizes without retraining.
We then train separate models on $768$- and $2048$-pixel
crops and evaluate each at the corresponding crop size,
using the same test locations.  

\section{METHOD}
\label{sec:method}

\begin{figure*}[t]
    \centering
    \includegraphics[width=\textwidth]{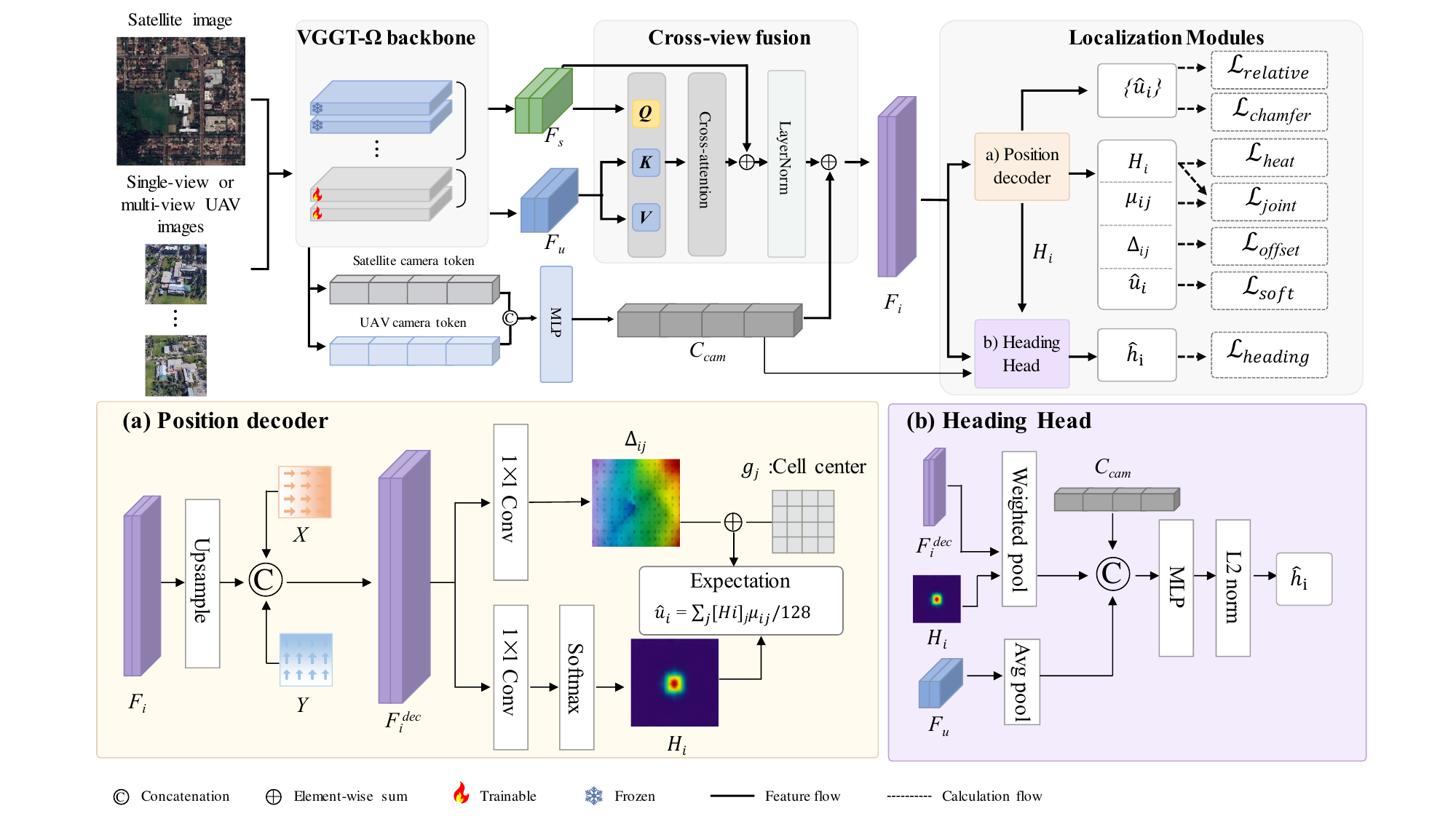}
    \vspace{-.3in}
    \caption{\textbf{Overview of \method{}.} The satellite image and
    multiple UAV views are jointly encoded by the VGGT-$\Omega$ backbone.
    Satellite-grid features query UAV visual features through cross-attention,
    after which a coordinate-aware decoder predicts a spatial distribution over
    continuous position hypotheses. The heading branch estimates a unit heading vector
for each UAV view.}
    \label{fig:model_pipeline}
    \vspace{-.15in}
\end{figure*}

As shown in Fig.~\ref{fig:model_pipeline}, \method{} first matches UAV visual
features to the satellite grid, then decodes the matched features into
continuous positions and headings.

\subsection{Geometry-grounded Localization}
\label{sec:geometry_grounding}

\noindent\textbf{Cross-view feature matching.}
We use a pretrained VGGT-$\Omega$ backbone \cite{wang2026vggtomega} to jointly
encode the satellite and UAV images. The resulting representations provide complementary visual and
geometry-aware information across the UAV views. To associate each
UAV view with a location in the map, satellite-grid features query the
corresponding UAV features through cross-attention. We fuse the matched
features with the original satellite features and a camera-conditioning
vector derived from the camera tokens of the satellite and UAV inputs. The fused representation
retains the satellite-grid layout and provides view-specific evidence for
localization.  

\noindent\textbf{Coordinate-aware decoding.}
Given the fused features, the decoder upsamples them and incorporates spatial
coordinate channels following CoordConv \cite{liu2018coordconv}. The coordinate channels give
each grid cell an explicit map position alongside its visual features.
For UAV view $i$, the decoder predicts a spatial probability map $H_i$ and
a two-dimensional offset at each cell. The heatmap is normalized by spatial
softmax, so the entries of $H_i$ sum to one.

\subsection{Position and Heading Modeling}
\label{sec:position_heading}

\noindent\textbf{Position estimation.}
Each cell represents a candidate UAV location. Adding its predicted offset
to the cell center gives a continuous position hypothesis
$\boldsymbol{\mu}_{ij}$, where $j$ indexes the output cells.
At inference, we select the hypothesis with the highest probability in $H_i$
and convert its output-grid coordinates to the input satellite crop,
obtaining $\hat{\mathbf p}_i$.
For differentiable position supervision, we use the expectation
$\mathbb E_{j\sim H_i}[\boldsymbol{\mu}_{ij}]$ during training.
We denote its crop-normalized form by $\hat{\mathbf u}_i$
and the corresponding normalized ground truth by $\mathbf u_i$.

\noindent\textbf{Heading estimation.}
The same probability map guides heading prediction. We use $H_i$ to pool
the localization features, giving greater weight to likely UAV locations.
The pooled features are combined with the camera-conditioning vector and
the globally pooled UAV features. An MLP maps this representation to a
two-dimensional vector, which is normalized to the unit heading vector
$\hat{\mathbf h}_i$. We recover the final bearing as
$\hat\theta_i=\operatorname{atan2}(\hat h_{\rm E},\hat h_{\rm N})$, using the
$(\emph{North},\emph{East})$ component ordering defined in
Section~\ref{sec:dataset_setting}.  

\noindent\textbf{Optimization objectives.}
To jointly train the heatmap and offsets, we give greater weight to
continuous position hypotheses near the ground truth. The target is
converted to output-grid coordinates, giving the joint position loss
\begin{equation}
\mathcal L_{\rm joint}
=-\log\!\left[
\sum_j [H_i]_j
\exp\!\left(-\frac{\|\boldsymbol{\mu}_{ij}-\mathbf p_i/4\|_2^2}{4.5}\right)
\right].
\label{eq:joint_position}
\end{equation}
Because the same distance term weights the heatmap score, a cell contributes
strongly only when its offset is also accurate, thereby encouraging sub-cell
precision at high-scoring locations. We additionally supervise the heatmap,
offsets, and expected position with $\mathcal L_{\rm heat}$,
$\mathcal L_{\rm offset}$, and $\mathcal L_{\rm soft}$, respectively.
These auxiliary terms stabilize position learning, and the combined objective is
\begin{equation}
\mathcal L_{\rm pos}
=
\mathcal L_{\rm joint}
+
\mathcal L_{\rm heat}
+
\mathcal L_{\rm offset}
+
\mathcal L_{\rm soft}.
\label{eq:localization_loss}
\end{equation}

For heading, we compare the predicted unit vector with the ground-truth
vector $\mathbf h_i$ using cosine distance:
\begin{equation}
\mathcal L_{\rm heading}=1-\hat{\mathbf h}_i^\top\mathbf h_i.
\label{eq:heading_loss}
\end{equation}
This representation avoids the discontinuity between $0^\circ$ and $360^\circ$. For multiple UAV views, we further supervise the trajectory using normalized
positions. The relative-displacement loss compares the predicted and
ground-truth displacement between consecutive views:
\begin{equation}
\mathcal L_{\rm relative}
=\operatorname{SmoothL1}\!\left(
\hat{\mathbf u}_{i+1}-\hat{\mathbf u}_i,
\mathbf u_{i+1}-\mathbf u_i
\right).
\label{eq:relative_loss}
\end{equation}
To also constrain the overall spatial layout, we apply symmetric Chamfer
distance between the predicted and ground-truth point sets:
\begin{equation}
\scalebox{0.9}{$\displaystyle
\mathcal L_{\rm chamfer}
=\frac{1}{2}\times\operatorname{mean}_i\!\left[
\min_j\|\hat{\mathbf u}_i-\mathbf u_j\|_2
+\min_j\|\mathbf u_i-\hat{\mathbf u}_j\|_2
\right].
$}
\label{eq:chamfer_loss}
\end{equation}
The mean and nearest-neighbor searches cover all views in the clip.
Unlike the relative-displacement loss, this term compares the point sets
without using their temporal order.

The final supervised objective combines the position, heading, and
trajectory terms:
\begin{equation}
\scalebox{0.85}{$\displaystyle
\mathcal L_{\rm total}
=
\mathcal L_{\rm pos}
+\lambda_{\rm head}\mathcal L_{\rm heading}
+\lambda_{\rm rel}\mathcal L_{\rm relative}
+\lambda_{\rm cham}\mathcal L_{\rm chamfer}
$}
\label{eq:full_loss}
\end{equation}
The position coefficient is fixed to one. Position and heading losses are
averaged over views, relative-displacement losses over consecutive pairs,
and the combined loss over the batch. The trajectory terms apply only when $T>1$.
Implementation and loss settings are detailed in
Section~\ref{sec:experiments}.

\section{EXPERIMENT}
\label{sec:experiments}

All experiments are conducted on \dataset{} using the
location-level train/validation/test split defined in
Section~\ref{sec:dataset_setting}.
Unless stated otherwise, the satellite crop is the default $630$~m reference.

\noindent\textbf{Implementation details and loss settings.}
We initialize the visual geometry backbone from the released VGGT-$\Omega$
1B-512 checkpoint \cite{wang2026vggtomega}. The last two pairs of frame and
inter-frame attention blocks, together with the proposed localization and
heading modules, remain trainable.
We train the full model with AdamW for 30 epochs, sampling 12,000
examples per epoch with a batch size of 1 and eight-step gradient
accumulation. The learning rates for the new heads and pretrained backbone are
$2\times10^{-4}$ and $5\times10^{-6}$, respectively, with a 5\% linear
warmup followed by cosine decay. During training, we sample the number of UAV views from
$T\in\{1,3,5,7,9\}$. Training takes about 40 hours on a 24-GB RTX 4090, and we
select the checkpoint based on validation localization and heading
performance. The default loss weights are
$\lambda_{\rm head}=0.25$, $\lambda_{\rm rel}=0.25$, and
$\lambda_{\rm cham}=0.10$. These coefficients are selected using the
validation locations only and are then fixed for all test-set evaluations.
The loss-weight study reported below is used only to characterize
sensitivity and is not used for hyperparameter selection. In the multi-view setting, we sample
UAV frames from the same
839 test locations and arrange them in temporal order.
Inputs with more views retain the frames used in inputs with
fewer views. We also evaluate a single-view reference using
the same sampling procedure.
Satellite and UAV inputs are resized to $512\times512$, and the decoder
outputs a $128\times128$ grid. In Eq.~\eqref{eq:joint_position},
$\mathbf p_i/4$ converts the input-crop target to output-grid coordinates;
the Gaussian bandwidth is $1.5$ cells, giving the denominator $4.5$.
The expected grid position is divided by $128$ to obtain $\hat{\mathbf u}_i$,
with target $\mathbf u_i=\mathbf p_i/512$. At inference, grid positions
are multiplied by four to recover input-crop coordinates. The cross-attended features are added to the satellite features and
layer-normalized before adding the broadcast camera-conditioning vector.
This vector is produced by an MLP from concatenated satellite and UAV camera
tokens. The decoder uses coordinate channels in $[-1,1]^2$ and
$0.5\tanh(\cdot)$ for offsets bounded within half a grid cell per component.
Auxiliary position losses consist of cross-entropy with a Gaussian heatmap target,
SmoothL1 regression of the offset at the ground-truth cell, and SmoothL1 regression
of the normalized expected position. We will also release the training and inference code, baseline adaptations,
and evaluation scripts to facilitate reproducibility. 

\begin{table}[t]
\centering

\scriptsize
\renewcommand{\arraystretch}{1.08}
\setlength{\tabcolsep}{1.0pt}
\resizebox{\columnwidth}{!}{
\begin{tabular}{l cc cccc}
\toprule
\multirow{2}{*}{\textbf{Method}} &
\multicolumn{2}{c}{\textbf{Localization Error (m)}} &
\multicolumn{4}{c}{\textbf{Localization Success (\%)}} \\
\cmidrule(lr){2-3}
\cmidrule(lr){4-7}
&
\textbf{Mean} &
\textbf{Median} &
\textbf{@10\,m} &
\textbf{@25\,m} &
\textbf{@50\,m} &
\textbf{@100\,m} \\
\midrule
\rowcolor{gray!20}
\multicolumn{7}{l}{\shortstack[l]{\textbf{Retrieval-based methods}}} \\
University-1652 baseline~\cite{zheng2020university1652}
& 230.72 & 221.13
& 0.09 & 0.65 & 2.71 & 11.88 \\
Sample4Geo~\cite{deuser2023sample4geo}
& 171.12 & 141.28
& 0.08 & 0.57 & 2.78 & 21.13 \\
MFRGN~\cite{wang2024mfrgn}
& 170.47 & 139.99
& 0.09 & 0.52 & 2.73 & 21.57 \\
DMNIL~\cite{chen2025dmnil}
& 165.24 & 134.95
& 0.09 & 0.50 & 2.54 & 21.84 \\
DenseUAV~\cite{dai2024denseuav}
& 163.63 & 141.69
& 0.06 & 0.32 & 1.42 & 14.95 \\
\midrule
\rowcolor{gray!20}
\multicolumn{7}{l}{\shortstack[l]{\textbf{Localization-based methods}}} \\
WildNav~\cite{gurgu2022wildnav}
& 264.58 & 263.56
& 0.01 & 0.02 & 0.07 & 0.22 \\
Reloc3r~\cite{dong2025reloc3r}
& 196.50 & 185.75
& 0.14 & 0.98 & 3.93 & 16.56 \\
Bearing-UAV~\cite{liu2026bearing}
& 197.90 & 178.69
& 0.10 & 0.59 & 2.66 & 15.90 \\
AnyVisLoc~\cite{ye2026anyvisloc}
& 164.15 & 137.04
& 0.07 & 0.44 & 2.14 & 20.33 \\
\midrule
VGGT-$\Omega$ zero-shot~\cite{wang2026vggtomega}
& 212.64 & 196.49
& 0.43 & 3.28 & 15.20 & 30.10 \\
VGGT-$\Omega$ fine-tuned~\cite{wang2026vggtomega}
& 88.15 & 69.17
& 5.93 & 22.43 & 40.58 & 61.74 \\
\textbf{\method{}}
& \textbf{82.33}
& \textbf{45.66}
& \textbf{17.34}
& \textbf{34.88}
& \textbf{52.08}
& \textbf{66.18} \\
\bottomrule
\end{tabular}
}
\caption{Single-view UAV--satellite metric localization on all 45,306
\dataset{} test frames. All methods use the same location-level split
and evaluation protocol; localization success is reported in percent.}
\label{tab:main_results}
\vspace{-.1in}
\end{table}


\begin{table}[tbp]
\centering

\renewcommand{\arraystretch}{1.08}
\setlength{\tabcolsep}{1.2pt}

\resizebox{\columnwidth}{!}{
\begin{tabular}{l|c|cc cc cccc}
\toprule
\multirow{2}{*}{\textbf{Method}} &
\multirow{2}{*}{\textbf{Views}} &
\multicolumn{2}{c}{\textbf{Position (m)}} &
\multicolumn{2}{c}{\textbf{Heading} (${\scriptstyle{}^\circ}$)} &
\multicolumn{4}{c}{\textbf{SR (\%)}} \\
\cmidrule(lr){3-4}
\cmidrule(lr){5-6}
\cmidrule(lr){7-10}
& &
\textbf{Mean} &
\textbf{Med.} &
\textbf{Mean} &
\textbf{Med.} &
\textbf{@10\,m} &
\textbf{@25\,m} &
\textbf{@50\,m} &
\textbf{@100\,m} \\
\midrule

VGGT-$\Omega$ zero-shot
& 3
& 270.98 & 232.94
& 41.04 & 27.61
& 0.40 & 3.54 & 15.53 & 30.43 \\

VGGT-$\Omega$ zero-shot
& 5
& 304.04 & 252.35
& 40.32 & 23.82
& 0.45 & 4.15 & 16.83 & 31.87 \\

VGGT-$\Omega$ zero-shot
& 7
& 311.26 & 251.32
& 39.04 & 22.76
& 0.73 & 4.58 & 18.81 & 33.92 \\

VGGT-$\Omega$ zero-shot
& 9
& 319.01 & 253.69
& 39.04 & 22.72
& 0.61 & 4.48 & 18.82 & 34.30 \\

\midrule

VGGT-$\Omega$ fine-tuned
& 3
& 83.00 & 66.77
& 38.23 & 31.82
& 6.16 & 23.32 & 40.52 & 63.37 \\

VGGT-$\Omega$ fine-tuned
& 5
& 80.21 & 66.83
& 37.40 & 29.53
& 6.82 & 25.79 & 42.36 & 64.24 \\

VGGT-$\Omega$ fine-tuned
& 7
& 77.39 & 62.83
& 36.43 & 26.40
& 6.66 & 26.73 & 44.07 & 66.56 \\

VGGT-$\Omega$ fine-tuned
& 9
& 76.68 & 61.73
& 36.39 & 25.43
& 6.70 & 26.94 & 44.48 & 66.55 \\

\midrule

\method{}
& 3
& 76.64 & 34.46
& 33.04 & 12.16
& 22.01 & 39.37 & 59.99 & 70.60 \\

\method{}
& 5
& 66.65 & 28.98
& 29.64 & 10.01
& 25.84 & 46.08 & 65.65 & 74.42 \\

\method{}
& 7
& 59.10 & 24.34
& 27.46 & 9.33
& 30.14 & 50.81 & 69.35 & 77.86 \\

\method{}
& 9
& \textbf{56.12}
& \textbf{21.96}
& \textbf{26.56}
& \textbf{8.73}
& \textbf{32.15}
& \textbf{53.26}
& \textbf{70.43}
& \textbf{78.65} \\

\bottomrule
\end{tabular}}
\caption{Localization and heading estimation in the multi-view
setting on \dataset{} for $T\in\{3,5,7,9\}$.
Localization success is reported in percent.}
\label{tab:multiview_results}
\vspace{-.1in}
\end{table}

\noindent\textbf{Compared methods.}
We compare \method{} with representative retrieval-based and direct-localization
approaches. The retrieval baselines are the University-1652 baseline
\cite{zheng2020university1652}, Sample4Geo \cite{deuser2023sample4geo},
MFRGN \cite{wang2024mfrgn}, DMNIL \cite{chen2025dmnil}, and DenseUAV
\cite{dai2024denseuav}. For direct localization, we evaluate WildNav
\cite{gurgu2022wildnav}, Reloc3r \cite{dong2025reloc3r},
Bearing-UAV \cite{liu2026bearing}, and AnyVisLoc \cite{ye2026anyvisloc}.
We evaluate VGGT-$\Omega$ \cite{wang2026vggtomega} both zero-shot and after
fine-tuning on our training split, to separate the gains of fine-tuning from
those of our localization modules. For the zero-shot VGGT-$\Omega$ baseline,
we use its stronger reference-plane projection: a robust reference plane is
estimated from the satellite branch, and the predicted UAV camera center is
projected onto this plane before being mapped to the satellite crop. This projection uses no test-set ground-truth correspondence or fitted
similarity transform.

For a consistent comparison, we evaluate all methods using the same
location-level train/validation/test split, satellite extent, coordinate
transformation, and metric evaluator. 
Except for the zero-shot VGGT-$\Omega$ baseline, all baselines are fine-tuned
on our training split. Retrieval methods extract overlapping satellite tiles of approximately
$100\times100$~m with an approximately $10$~m stride along each axis
($55\times55=3025$ candidates for the default crop) and return the
center of the top-ranked tile, whereas direct-localization methods retain continuous position
outputs, which we map into the same metric coordinate frame.  

\subsection{Comparison with the State-of-the-Art Methods}
\label{sec:main_comparison}

We first evaluate single-view localization over all 45,306 test frames and
report the results in Table~\ref{tab:main_results}. As shown in Table~\ref{tab:main_results},  
\method{} achieves a mean error of $82.33$~m and a median error
of $45.66$~m, with localization success rates of $17.34\%$, $34.88\%$,
$52.08\%$, and $66.18\%$ at 10, 25, 50, and 100~m, respectively. The fine-tuned
VGGT-$\Omega$ baseline provides the strongest comparison, with a mean error of
$88.15$~m and a median error of $69.17$~m. Relative to the fine-tuned VGGT-$\Omega$ baseline, \method{} reduces the
median position error by $34.0\%$, from $69.17$~m to $45.66$~m, while also
lowering the mean error from $88.15$~m to $82.33$~m.

To understand this gap, we consider the two baseline families. Retrieval
accuracy depends on cross-view matching and gallery quantization. Direct
localization removes the discrete output constraint, but still faces strong
appearance and geometric differences between oblique UAV views and satellite
imagery. For example, AnyVisLoc yields $164.15/137.04$~m mean/median error,
compared with $82.33/45.66$~m for \method{}, indicating that continuous
prediction alone does not resolve these cross-view differences. Zero-shot VGGT-$\Omega$ yields mean/median position errors
of $212.64/196.49$~m. Fine-tuning substantially reduces these
errors, and the proposed localization modules provide
further improvements.


\subsection{Ablation Studies and Further Discussion}
\label{sec:ablation}

\noindent\textbf{Multi-view setting.}
Table~\ref{tab:main_results} reports the canonical single-view evaluation
over all 45,306 test frames. For the controlled multi-view study, we instead
construct nested frame sets using the same temporal sampling procedure for
all view counts. We therefore report a separate $T=1$ reference under this
sampling protocol, which gives mean/median position errors of
$212.38/194.10$~m for zero-shot VGGT-$\Omega$ and
$87.07/41.85$~m for \method{}. We then evaluate \method{} with increasing numbers of UAV views.
Table~\ref{tab:multiview_results} reports results for
$T\in\{3,5,7,9\}$. Under this controlled sampling protocol, increasing the number of views
from one to nine reduces the mean position error from $87.07$~m to
$56.12$~m and the median position error from $41.85$~m to $21.96$~m.
SR@10~m increases from $19.31\%$ to $32.15\%$, while the
median heading error decreases from $16.47^\circ$ to
$8.73^\circ$. Thus, additional UAV views improve localization accuracy. The gains from additional UAV views diminish beyond five views,
particularly in median heading error. The zero-shot VGGT-$\Omega$
baseline does not improve consistently as views are added, while its
fine-tuned counterpart remains less accurate than \method{} for
$T\in\{3,5,7,9\}$.

\noindent\textbf{Effect of localization components.}
We next study how coordinate channels, the joint heatmap--offset objective,
and heading supervision affect single-view localization through the component
ablations in Table~\ref{tab:component_ablation}.
Across the evaluated configurations, the full model achieves the lowest mean
position error and the highest localization success rates. The
coordinate-channel-only variant has a nearly identical median error
($45.56$~m vs.\ $45.66$~m), while removing heading supervision changes
SR@10~m only slightly ($17.08\%$ vs.\ $17.34\%$).
With coordinate channels and heading supervision
fixed, adding the joint position objective in Eq.~\eqref{eq:joint_position} to the
auxiliary heatmap, offset, and expected-position objectives reduces mean/median
position error from $82.80/47.85$~m to $82.33/45.66$~m and increases SR@10~m
from $15.97\%$ to $17.34\%$. This controlled comparison is consistent with coupling  
spatial probability and sub-cell refinement within a single continuous-position
objective.

\begin{table}[tbp]
\centering

\scriptsize
\renewcommand{\arraystretch}{1.08}
\setlength{\tabcolsep}{1.0pt}

\resizebox{\columnwidth}{!}{
\begin{tabular}{l|ccc|cc cc}
\toprule
\textbf{Configuration} &
\textbf{XY} &
\textbf{Joint H--O} &
\textbf{Heading} &
\textbf{Mean (m)} &
\textbf{Median (m)} &
\textbf{@10\,m} &
\textbf{@100\,m} \\
\midrule

CoordConv only
& $\checkmark$ & &
& 83.30 & \textbf{45.56} & 15.97 & 65.87 \\

Joint H--O only
& & $\checkmark$ &
& 84.81 & 49.20 & 16.45 & 65.40 \\

Heading only
& & & $\checkmark$
& 85.64 & 47.20 & 15.19 & 64.94 \\

\midrule

\textit{w/o} CoordConv
& & $\checkmark$ & $\checkmark$
& 87.15 & 50.15 & 16.57 & 64.15 \\

\textit{w/o} Joint H--O
& $\checkmark$ & & $\checkmark$
& 82.80 & 47.85 & 15.97 & 65.81 \\

\textit{w/o} Heading
& $\checkmark$ & $\checkmark$ &
& 84.71 & 48.66 & 17.08 & 65.32 \\

\midrule

Full \method{}
& $\checkmark$ & $\checkmark$ & $\checkmark$
& \textbf{82.33} &
45.66 &
\textbf{17.34} &
\textbf{66.18} \\

\bottomrule
\end{tabular}
}
\caption{Component ablation on all 45,306 \dataset{} test frames with
single-view inputs. XY denotes coordinate channels, and Joint H--O
denotes the joint heatmap--offset objective.}
\label{tab:component_ablation}
\end{table}

\noindent\textbf{Sensitivity to loss weights.}
We further study the sensitivity of \method{} to the three loss weights in
Eq.~\eqref{eq:full_loss}. Starting from the fixed default configuration, we
vary one coefficient at a time while keeping the position-loss coefficient
equal to one and the remaining coefficients unchanged. Each configuration
uses the same training procedure with $T\in\{1,3,5,7,9\}$, and we report its
single-view performance on the 45,306 test frames in
Table~\ref{tab:coefficient_ablation}. Increasing $\lambda_{\rm rel}$ from $0.25$ to $0.30$ or $0.35$ increases
the mean position error from $82.33$~m to $93.53$ and $93.21$~m,
respectively. In contrast, varying $\lambda_{\rm cham}$ over the tested range
changes the mean position error only moderately, from $82.33$ to $84.03$~m.
The median heading error is $20.57^\circ$ at
$\lambda_{\rm cham}=0.08$ and $20.81^\circ$ at the default value of $0.10$.
We therefore regard the small differences among nearby configurations as
sensitivity trends rather than precise rankings.

\begin{table}[tbp]
\centering
\renewcommand{\arraystretch}{1.08}
\setlength{\tabcolsep}{2.8pt}
\scriptsize
\resizebox{0.9\columnwidth}{!}{
\begin{tabular}{ccc| cc c}
\toprule
\multicolumn{3}{c}{\textbf{Loss Weight}} &
\multicolumn{2}{c}{\textbf{Position (m)}} &
\textbf{Heading (${}^\circ$)} \\
\cmidrule(lr){1-3}
\cmidrule(lr){4-5}
$\lambda_{\rm head}$ &
$\lambda_{\rm rel}$ &
$\lambda_{\rm cham}$ &
\textbf{Mean} &
\textbf{Median} &
\textbf{Median} \\
\midrule

0.15 & 0.25 & 0.10 & 85.95 & 48.85 & 21.45 \\
0.20 & 0.25 & 0.10 & 85.07 & 48.52 & 21.04 \\
0.25 & 0.25 & 0.10 & 82.33 & 45.66 & 20.81 \\
0.30 & 0.25 & 0.10 & 88.48 & 51.14 & 22.57 \\
0.35 & 0.25 & 0.10 & 84.53 & 49.27 & 21.63 \\

\midrule

0.25 & 0.15 & 0.10 & 85.07 & 47.91 & 21.14 \\
0.25 & 0.20 & 0.10 & 85.96 & 49.64 & 21.55 \\
0.25 & 0.25 & 0.10 & 82.33 & 45.66 & 20.81 \\
0.25 & 0.30 & 0.10 & 93.53 & 58.94 & 26.15 \\
0.25 & 0.35 & 0.10 & 93.21 & 59.06 & 25.69 \\

\midrule

0.25 & 0.25 & 0.06 & 83.44 & 47.34 & 21.32 \\
0.25 & 0.25 & 0.08 & 84.03 & 46.05 & 20.57 \\
0.25 & 0.25 & 0.10 & 82.33 & 45.66 & 20.81 \\
0.25 & 0.25 & 0.12 & 83.55 & 47.05 & 21.58 \\
0.25 & 0.25 & 0.14 & 83.12 & 47.91 & 20.79 \\

\bottomrule
\end{tabular}}
\caption{Loss-weight ablation on all 45,306 \dataset{} test frames with
single-view inputs. The position-loss coefficient is fixed to one;
one coefficient is varied at a time while the others remain at their
default values.}
\label{tab:coefficient_ablation}
\vspace{-.1in}
\end{table}

\begin{figure*}[t]
    \centering
    \vspace{-.15in}
    \includegraphics[width=\textwidth]{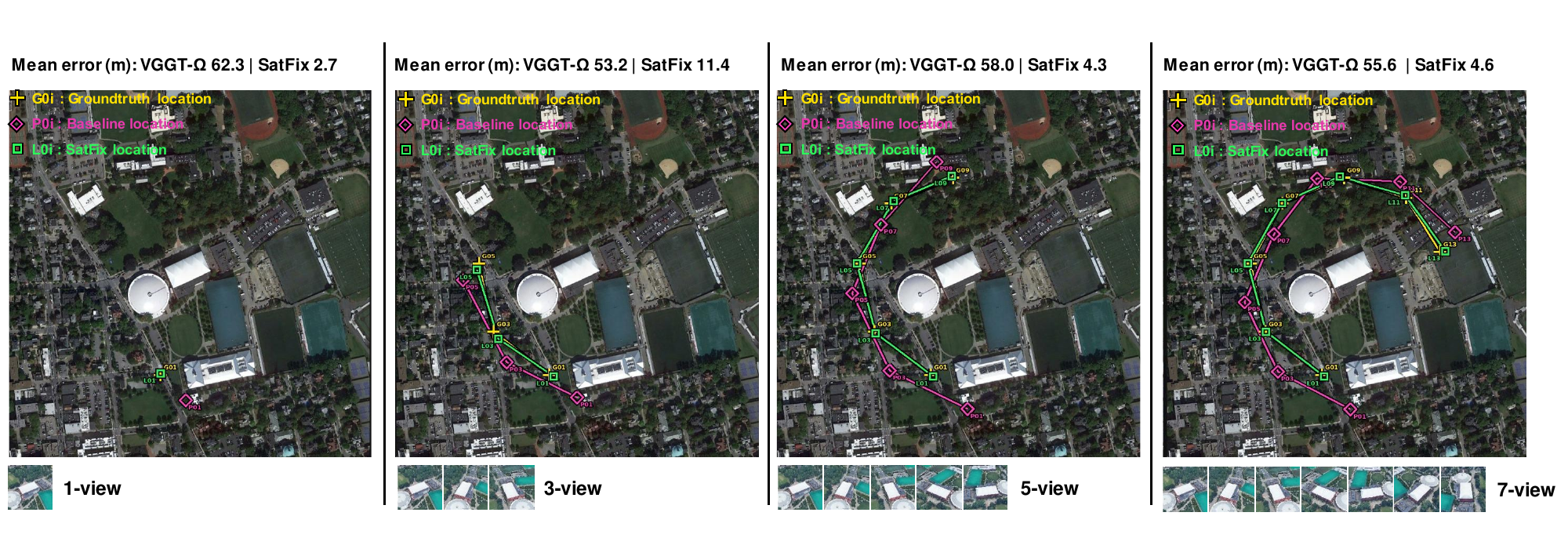}
     \vspace{-.35in}
\caption{Qualitative comparison of multi-view UAV--satellite
localization on \dataset{}. From one to seven UAV views, ground-truth locations
are shown in yellow, predictions from the VGGT-$\Omega$ zero-shot baseline
in magenta, and \method{} predictions in green. The UAV views used
in each case are shown below the corresponding satellite map. The reported mean error is averaged over the displayed localization points
within each example and is used only to summarize the qualitative result.}
    \label{fig:qualitative_multiview}
    \vspace{-.1in}
\end{figure*}

\noindent\textbf{Effect of satellite extent.}
Beyond model components, we examine the effect of the satellite extent
(Table~\ref{tab:satellite_extent}). We vary the physical extent of the map while
keeping the network input fixed at $512\times512$. Without retraining,
reducing the satellite extent from the default $630$~m setting to $473$~m
decreases mean error from $82.33$~m to $63.67$~m. Increasing the extent to
approximately $1261$~m raises mean error to $199.15$~m and reduces
SR@50~m to $1.55\%$. To test whether training mitigates this performance loss, we also train and
evaluate on matching crop sizes. Training on $2048$-pixel crops reduces
mean error from $199.15$~m to $97.66$~m. In contrast, on $768$-pixel test
crops, training on the same crop size does not outperform the model trained
on $1024$-pixel crops.

\noindent\textbf{Qualitative comparison.}
Figure~\ref{fig:qualitative_multiview} visualizes localization with one,
three, five, and seven UAV views. The satellite maps compare
ground-truth UAV locations (yellow) with predictions from zero-shot
VGGT-$\Omega$ (magenta) and \method{} (green), with the corresponding
UAV inputs shown below.

\noindent\textbf{Scope and limitations.}
The median position error of $21.96$~m and median heading
error of $8.73^\circ$ at $T=9$ suggest potential for regional
position correction.
The present experiments evaluate localization accuracy, not closed-loop
navigation, integration with odometry, or UAV--UGV cooperative routing
\cite{mondal2025riskaware}. Position errors above $100$~m remain in $33.82\%$ of single-view
predictions and $21.35\%$ at $T=9$. The evaluation assumes satellite coverage of the operating
region, and robustness to map aging, seasonal changes, sensor
shift, and incomplete maps remains to be tested. We further assume a coarse prior on the operating region, which bounds the search space to the provided crop; the heatmap probability is not calibrated, so the model does not yet signal when its estimate should be discarded; and the reported latency is measured on a desktop GPU, while deployment efficiency
on embedded hardware remains to be evaluated.  

\noindent\textbf{Inference time.}
We measure the complete single-view inference pipeline excluding only the
one-time model loading cost. The timing includes input preprocessing,
satellite/UAV encoding, localization, and coordinate decoding.
We compare against four direct-localization baselines under the same timing
scope. Bearing-UAV and AnyVisLoc take $0.037$ and $0.052$~s per query,
respectively, while WildNav and Reloc3r take $0.250$ and $0.382$~s.
\method{} takes less than $0.1$~s per query and achieves lower mean and
median position errors than these four baselines
(Table~\ref{tab:main_results}). Although slower than Bearing-UAV and
AnyVisLoc, \method{} is faster than WildNav and Reloc3r.  

\begin{table}[tbp]
\centering

\setlength{\tabcolsep}{4pt}
\renewcommand{\arraystretch}{1.15}
\resizebox{0.95\columnwidth}{!}{
\small
\begin{tabular}{cc c ccc}
\toprule
\textbf{Train Crop} &
\textbf{Test Crop} &
\textbf{Extent (m)} &
\textbf{Mean (m)} &
\textbf{Median (m)} &
\textbf{SR@50\,m (\%)} \\
\midrule

1024 & 768
& $\sim$473
& 63.67 & 36.94 & 62.57 \\

1024 & 1024
& $\sim$630
& 82.33 & 45.66 & 52.08 \\

1024 & 2048
& $\sim$1261
& 199.15 & 183.99 & 1.55 \\

\midrule

768 & 768
& $\sim$473
& 99.94 & 59.75 & 43.69 \\

2048 & 2048
& $\sim$1261
& 97.66 & 85.24 & 15.79 \\

\bottomrule
\end{tabular}}
\caption{Effect of the satellite extent on \dataset{}.
Train Crop and Test Crop are the pixel sizes of the square satellite crop;
Extent is the corresponding physical side length in meters. All crops are
resized to $512\times512$.} 
\label{tab:satellite_extent}
\vspace{-.1in}
\end{table}

\section{CONCLUSION}
\label{sec:conclusion}

We presented \method{} for absolute metric UAV localization from one or a few
oblique UAV views within a geo-referenced satellite region. We also introduced
\dataset{}, a benchmark derived from University-1652 with a location-level
train/validation/test split, enlarged regional satellite references, and
continuous position and viewing-heading annotations.
On \dataset{}, \method{} grounds feed-forward geometric representations
to the satellite coordinate frame and jointly estimates continuous
position and heading using a single model for both single-view and multi-view inputs. Across the reported multi-view settings, \method{} achieves lower position
and heading errors than both zero-shot and fine-tuned VGGT-$\Omega$ baselines,
reaching $56.12/21.96$~m mean/median position error and an
$8.73^\circ$ median heading error with nine views.


\bibliographystyle{IEEEtran}
\bibliography{IEEEexample}

\end{document}